\documentclass[acmtog]{acmart}
\usepackage{listings,xcolor}

\AtBeginDocument{%
  \providecommand\BibTeX{{%
    \normalfont B\kern-0.5em{\scshape i\kern-0.25em b}\kern-0.8em\TeX}}}

\setcopyright{none}
\acmYear{2022}

\acmBooktitle{} 
\acmPrice{}
\acmISBN{}

\usepackage{color}
\usepackage{soul}
\usepackage{multirow}
\usepackage{rotating}
\usepackage{amsmath}
\usepackage{algorithmicx}
\usepackage{wrapfig}
\usepackage[normalem]{ulem}
\usepackage{booktabs}
\usepackage{wrapfig}

\definecolor{turquoise}{cmyk}{0.65,0,0.1,0.1}
\definecolor{purple}{rgb}{0.65,0,0.65}
\definecolor{darkgreen}{rgb}{0.0, 0.5, 0.0}
\definecolor{darkred}{rgb}{0.5, 0.0, 0.0}
\definecolor{darkblue}{rgb}{0.0, 0.0, 0.5}
\definecolor{blue}{rgb}{0.0, 0.0, 1.0}

\newcommand{\erase}[1]{}

\newcommand{\hide}[1]{{}}

\definecolor{darkblue}{rgb}{0.0, 0.2, 0.5}

\usepackage{marginnote}

\begin{document}


\title{Co-design of Embodied Neural Intelligence via Constrained Evolution}
\author{Zhiquan Wang}
\email{wang4490@purdue.edu}
\orcid{}
\affiliation{%
  \institution{Purdue University}
  \streetaddress{305 N University St.}
  \city{West Lafayette, }
  \state{Indiana}
  \country{USA}
  \postcode{47907-2021}
}

\author{Bedrich Benes}
\email{bbenes@purdue.edu}
\orcid{0000-0002-5293-2112}
\affiliation{%
  \institution{Purdue University}
  \streetaddress{305 N University St.}
  \city{West Lafayette, }
  \state{Indiana}
  \country{USA}
  \postcode{47907-2021}
}

\author{Ahmed H. Qureshi}
\email{qureshi7@purdue.edu}
\orcid{}
\affiliation{%
  \institution{Purdue University}
  \streetaddress{305 N University St.}
  \city{West Lafayette, }
  \state{Indiana}
  \country{USA}
  \postcode{47907-2021}
}

\author{Christos Mousas}
\email{cmousas@purdue.edu}
\orcid{}
\affiliation{%
  \institution{Purdue University}
  \streetaddress{Knoy Hall of Technology}
  \city{West Lafayette, }
  \state{Indiana}
  \country{USA}
  \postcode{47907-2021}
}

\renewcommand{\shortauthors}{Wang, et al.}

\begin{abstract}
We introduce a novel co-design method for autonomous moving agents' shape attributes and locomotion by combining deep reinforcement learning and evolution with user control. Our main inspiration comes from evolution, which has led to wide variability and adaptation in Nature and has the potential to significantly improve design and behavior simultaneously. Our method takes an input agent with optional simple constraints such as leg parts that should not evolve or allowed ranges of changes. It uses physics-based simulation to determine its locomotion and finds a behavior policy for the input design, later used as a baseline for comparison. The agent is then randomly modified within the allowed ranges creating a new generation of several hundred agents. The generation is trained by transferring the previous policy, which significantly speeds up the training. The best-performing agents are selected, and a new generation is formed using their crossover and mutations. The next generations are then trained until satisfactory results are reached. We show a wide variety of evolved agents, and our results show that even with only 10\% of changes, the overall performance of the evolved agents improves 50\%. If more significant changes to the initial design are allowed, our experiments' performance improves even more to 150\%. Contrary to related work, our co-design works on a single GPU and provides satisfactory results by training thousands of agents within one hour.  
\end{abstract}



\begin{teaserfigure}
   \includegraphics[width=\textwidth]{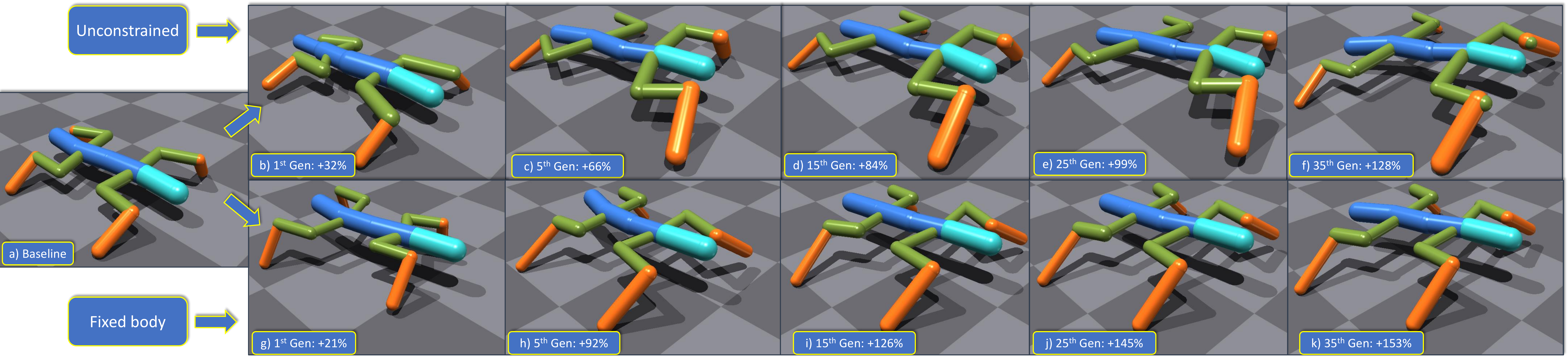}
   \caption{User-created agent is trained to walk with the state-of-the art PPO training (a). Top row: The agent is evolved to complete the same task without any constraints. Its morphology remains the same, but the evolutionary algorithm changes its parameters to perform the task better. The reward function value of the baseline agent is 100\% and it improves through the evolution to 132, 166, 184, 199, and 228\%. Bottom row: The body of the agent is restricted and cannot change through the evolution. Only the legs can evolve. The reward improves to 121, 192, 226, 245, and 253\% of the baseline design.}
   \label{fig:teaser}
\end{teaserfigure}

\maketitle

\section{Introduction}
Creating autonomous moving agents (e.g., robots, creatures) is a significant open problem with applications ranging from robotics to animation~\cite{todorov2012mujoco}. Their manual creation and motion design offer a high level of control but do not scale and are prone to errors. Automatic generation does not always lead to desired morphology and topology. Moreover, having the agents react to the environment requires the design of behavioral policies. Recent approaches focused on the automatic design of behavior policies, and significant advances have been achieved with the help of deep reinforcement learning (DeepRL) combined with motion simulation and fine-designed reward/objective function in physics-based environments~\cite{haarnoja2018soft,schulman2015trust,schulman2015high}.

While a large body of related work has addressed virtual agent behavior and control policy design, the co-design of a virtual agent shape and its corresponding control policy is an open research problem. While structural and behavioral co-design is the natural way for living forms, it is a challenging computational problem because the search space is ample and changes with each new agent's configuration. Existing algorithms optimizing the agent and its controller either use simple configurations (e.g., 2D space, voxels)~\cite{bhatia2021evolution}, or they often lead to structures that deviate from the initial design considerably. However, it is essential to balance the optimized structure, and the initial structure as the uncontrolled optimization may lead to a significantly different shape from the user's expectations. However, it is not necessary to optimize the agent by exploring different structures as the subtle changes of the initial design can increase its performance. We need a new optimization method to search space efficiently and constrain the morphology within the designer's expectations. 

Our first key observation comes from evolutionary algorithms that address the wide variability of forms and their adaptation~\cite{pfeifer2006body}. Recent advancements in DeepRL have provided us with ways to learn a single, universal behavior policy for a wide range of physical structures, resulting in less memory footprint and efficient behavior learning in large-scale settings~\cite{gupta2022metamorph}. Therefore, using universal DeepRL frameworks have the potential to provide an efficient way to explore the large solution space and design evolution-based methods. Our second key observation comes from the high variation that the evolutionary design often causes. This is often undesirable and providing user constraints over the way the agents evolve has the potential to control the agent's shape and prune the search space significantly. 

We propose a novel evolution-based method that can optimize the 3D physical parameters of an agent and its corresponding controller simultaneously within a user-defined boundary. Our work aims to generate various agents with similar physics attributes within the range of user inputs and a universal controller for them to walk in the given environment. The user input defines the range of the body part's length, radius, and range of joints' angle affecting the agents' kinematic and physics attributes. Our evolution-based method creates new agents based on the user-given template agent and optimizes their performance by generation. For each generation, we perform a training phase first to train a policy net with Proximal Policy Optimization (PPO) to control agents' motion in this generation. Our method builds on the recent work of \citeauthor{gupta2022metamorph}~\shortcite{gupta2022metamorph} that allows for learning of a universal controller over a modular robot design space. We designed a Multiple Layer Perceptron (MLP)/multi-head self-attention-based policy that can control all the agents with a single deep neural network. After the training phase, we measure the agents' performance and create a new generation by selecting high-performance agents and merging their attributes represented as genes. Through this evolution, we could quickly produce agents with high performance with several generations and achieve performance much higher than randomly generated agents higher than the template agent. The user controls what and how much can be modified through evolution, leading to agents that vary slightly from the original design but achieve significantly better performance. An example in Fig.~\ref{fig:teaser} shows the original design (a) and its performance. When the body changes are not allowed, our algorithm evolves a new, better-performing agent (b). Enabling the body modifications improves the performance even more (c), and allowing mutations causes more significant alterations to the original design, increasing the performance even more (d). The same agent then evolves while its body shape is fixed (g-k). 

We claim the following contributions: 1) an evolution-based optimization that produces agent structure that holds the design requirement and fits the given task, 2) our method is fast as we train one generation of agents at a time instead of a single agent 2) a universal policy can control various agents for a specific task, and 3) user control over the allowable agent's modifications. 

\begin{figure*}[hbt]
\centering
\includegraphics[width=0.99\linewidth]{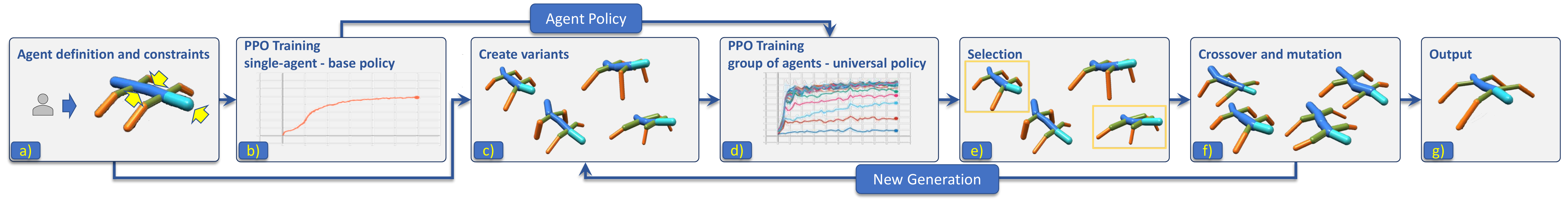}
  \caption{\textbf{Overview:} The agent is either generated randomly or with user support. The user also defined constraints (yellow arrows) (a). The initial Proximal Policy Optimization (PPO) trains the input agent to provide baseline agent policy (b). The system then creates variants of the initial model (c) and trains them all together with universal PPO (d). Selection (e), crossover, and mutation (f) create a new generation that is trained again. The system outputs the best(s) co-designed agents and their policies (g).}
  \label{fig:overview}\vspace{-3mm}
\end{figure*}
\section{Related Work}
We related our work to procedural modeling, physics-based animation, (deep) reinforcement learning for agent motion synthesis, and co-design of structure and behavior. 

\textbf{Procedural models} generate a model by executing a code, and the procedural rules and their parameters define a class of generated objects. \textbf{Procedural animation} automatically generates animation sequences that provide a diverse series of actions than could otherwise be created using predefined motion clips. A seminal example is the work of \citeauthor{reynolds1987flocks}~\shortcite{reynolds1987flocks} who introduced a simple reactive control of procedural agents that faithfully recreates complex motion patterns of flocks of birds and schools of fish. Similar approaches have been widely applied to crowd simulation (e.g.,~\cite{thalmann2012crowd,braun2003modeling,narain2009aggregate}). However, procedural animation is unsuitable for low-level agent control, and its common problem is the low level of control.

\textbf{Physics-based animation} represents the agents as interconnected rigid bodies with mass and moment of inertia controlled by joint torques or muscle models~\cite{won2021control}. As the control mechanism of an agent greatly affects the motion quality, the choice of control method is important depending on the task. \citeauthor{peng2017learning}~\shortcite{peng2017learning} compared the difference across torque control, PD controller, and muscle base control. 

Depending on an appropriate control method, many methods work on the control policy to synthesize realistic locomotion. One approach utilizes motion equations or implicit constraints to optimize the locomotion generated physics-based gaits by numerically integrating equations of motion~\cite{raibert1991animation}. \citeauthor{van1994virtual}~\shortcite{van1994virtual} developed a periodic control method with cyclic control graph~\cite{mordatch2012discovery} that applies a contact-invariant optimization to produce symmetry and periodicity fully automatically. The design of a physics-based controller remains challenging as it relies on the appropriate design of the agent and the task-specific objective functions assigned to it. 

An alternative approach is learning to synthesize motions from a motion dataset or reference motion clips~\cite{liu2005learning, yin2007simbicon, liu2017learning, chentanez2018physics, won2020scalable, won2021control}. One example is the real-time interactive controller based on human motion data that predicts the forces in a short window~\cite{da2008simulation} and the simulation of a 3D full-body biped locomotion by modulating continuously and seamlessly trajectory~\cite{lee2010data}. \citeauthor{wampler2014generalizing}~\shortcite{wampler2014generalizing} applied joint inverse optimization to learn the motion style from the database.

\textbf{Deep Reinforcement Learning (DeepRL)} provides a control policy for agents automatically. Deep reinforcement has been proven effective in diverse, challenging tasks, such as using a finite state machine (FSM) to guide the learning target of RL and drives a 2D biped walking on different terrains~\cite{peng2015dynamic}.  \citeauthor{yu2018learning}~\shortcite{yu2018learning} encouraged low-energy and symmetric motions in loss functions, and \citeauthor{abdolhosseini2019learning}~\shortcite{abdolhosseini2019learning} address the symmetry from the structure of policy network, data duplication, and loss function and they also handle different types of robots or terrains. One of the drawbacks is the loss of direct control of the learning target because the implicit function does not provide a clear learning target for the agent. Combining motion data has the potential to address this issue by giving an imitation target. With the assistance of motion reference, the learning process can discard huge meaningless motion and dramatically reduce the exploration of action space. \citeauthor{peng2018deepmimic}~\shortcite{peng2018deepmimic} enables the learning of challenging motion tasks by imitating motion data or video frames directly~\cite{peng2018sfv}. \citeauthor{won2019learning}~\shortcite{won2019learning} handle different shape variations of a virtual character. However, the learning from the unstructured motion database or the inaccuracy in the motion reference can make the learning of policy difficult. A fully automated approach based on adversarial imitation learning was introduced in~\cite{peng2021amp} to address this problem by generating new motion clips. Recently, \citeauthor{peng2022ase}~\shortcite{peng2022ase} combined adversarial imitation learning and unsupervised RL techniques to develop skill embeddings that produce life-like behaviors for virtual characters. The characters learn versatile and reusable physically simulated skills. One limitation of~\cite{peng2021amp, peng2022ase} is the need for a well-designed character in terms of density, length, and joint properties to perform the given task. Our work addresses this problem by combining RL and evolution.

\textbf{Co-optimizing design and behavior} attempts to optimize behavior or function and shape simultaneously. The seminal work of \citeauthor{sims1994evolving}~\shortcite{sims1994evolving} uses genetic algorithms~\cite{holland,koza95} to evolve 3D creatures by using physics-based simulation, neural networks, genetic algorithms, and competition. Evolution has also been used to design the shape of robots~\cite{bongard2013evolutionary,ha2019} and neural graph evolution has been applied to design robots in~\cite{wang2019neural}. Our work is inspired by the recent work (RoboGrammar)~\cite{zhao2020robogrammar} that uses graph search to optimize procedural robots for various terrains. RoboGrammar uses a set of well-tuned fixed body attributes (length, density, control parameters), while our method evolves the body attributes of the virtual agents. Close to our work is~\cite{bhatia2021evolution} that uses co-design via evolution to co-optimizing the design and control of 2D grid-based soft robots. This method works in 2D on a fixed set of agent parts and trains each agent individually, while our approach uses group training that significantly shortens training. This is inspired by \citeauthor{gupta2022metamorph}~\shortcite{gupta2022metamorph}, which controls different agents with one universal controller. We designed our universal controller with an MLP network instead of the self-attention layer as it is faster to train and provides similar results. Our controller handles agents with the same topology but different body attributes. The second work~\cite{gupta2021embodied} evolves the agent's structure by mutations and sampling without merging the parents' genes to reproduce the children and does not provide freedom of control over the agent's design during evolution.

\section{Overview}\label{sec:overview}
The input to our method (see Fig.~\ref{fig:overview} a) is an agent that was either provided by the user or generated randomly. The user can also define constraints that guide the changes in the agent form. Examples of the constraints (marked schematically as yellow arrows) are the ranges of the allowed changes in the length of the body, the width of legs, etc. Our method improves the performance of the physically simulated agent within the constraints via evolution and ensures the result does not deviate from users' expectations. The constraints do not need to be tuned carefully.

The input agent is trained (Fig.~\ref{fig:overview} b) by the Proximal Policy Optimization (PPO) in a physics-based environment as a simulated robot with a rigid body, collision detection, shape, and motors to perform a task. The output of this training is used as a baseline for evaluating the performance of the following stages of the algorithm. The learned policy is transferred into the agent's generation (Fig.~\ref{fig:overview} d) as a start policy that accelerates the following generations' training with encoded motion prior. 

The algorithm then enters into the co-design phase of evolution. The system creates several hundreds of variants of the agent by randomly sampling the allowed ranges of the parameters of the input agent (Fig.~\ref{fig:overview} c). The initial generation of agents is trained with the universal PPO, which significantly accelerates the training time and allows training on a single GPU. The trained agents are sorted according to their fitness, and the top agents are selected (Fig.~\ref{fig:overview} e). The selected agents undergo crossover and mutation to generate a new generation (Fig.~\ref{fig:overview} f), and the new generation is trained by bootstrapping with the policy from the parent generation. During the evolution, the agent keeps improving their attributes. The entire algorithm stops either when the improvement is insignificant or when the user decides that the output is satisfactory.

\section{Agent Description}\label{sec:descr}
Our agent description can be used in DeepRL frameworks, flexible supports physics-based simulation, and allows for a fast definition or user constraints. 

\subsection{Shape}\label{sec:shape}
\begin{figure}[hbt]
\centering
\includegraphics[width=.8\linewidth]{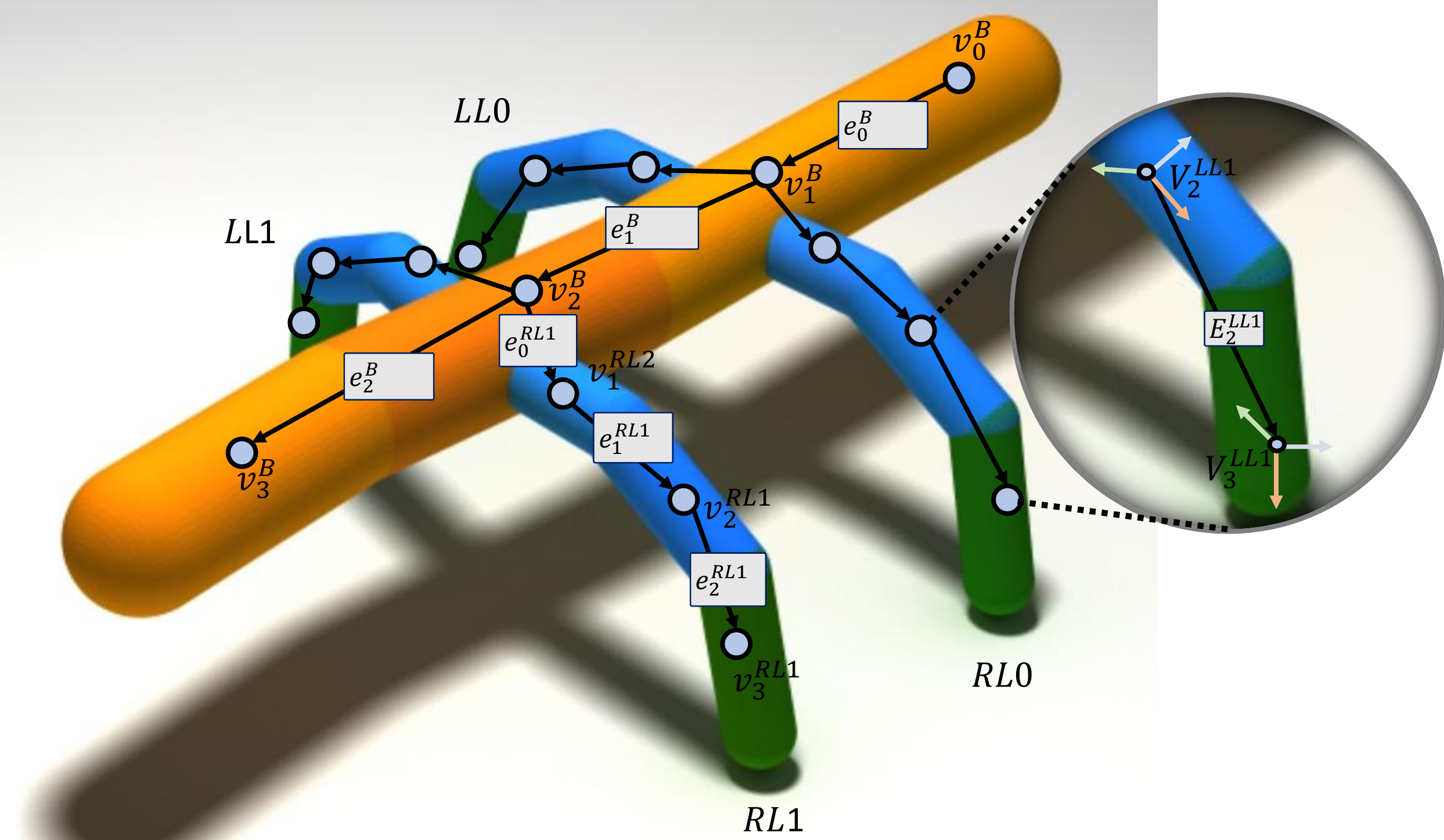}
  \caption{An example of an agent, its corresponding topological graph, and the coordinate systems of the joints (inset).}\vspace{-3mm}
  \label{fig:agent}
\end{figure}
The agent is represented as a directed acyclic graph $\mathcal{G}=\{V,E\}$ with vertices $v_i\in{V}$ and edges $e_{i,j}: v_i\rightarrow{v_j}$. Each $v_j$ corresponds to a node that connect different parts of the agents and each $e_{ij}$ is a joint that corresponds to connecting two parts  (nodes $v_i$ and $v_j$) of the agent's body (see Fig.~\ref{fig:agent}).

Each agent consists of two building blocks: body parts are denoted by the upper index~$B$, and legs with the foot are denoted by~$LL$ and~$RL$ for the left and right leg, respectively. The acyclic graph is a tree with the root being always the node~$v_0^{B}$. An example in Fig.~\ref{fig:agent} shows an agent with two pairs of legs and a body with four body parts. An additional index further distinguishes each leg, e.g., the third vertex on the second left leg from the torso has index~$v_1^{LL2}$ (indexed from zero).

While the topology of the agent is described by the graph~$\mathcal{G}$, the geometry is captured by additional data stored in each graph vertex~$v$ that is called agent's \textit{attributes}. Each body part is represented as a generalized cylinder (a capsule), and we store its local coordinate system, orientation, radius, and length. The edges also store the rotation axis and rotation range. 

\begin{wrapfigure}[7]{r}[0pt]{0.6\linewidth}
\centering
    \vspace{-6mm}
    \includegraphics[width=1\linewidth]{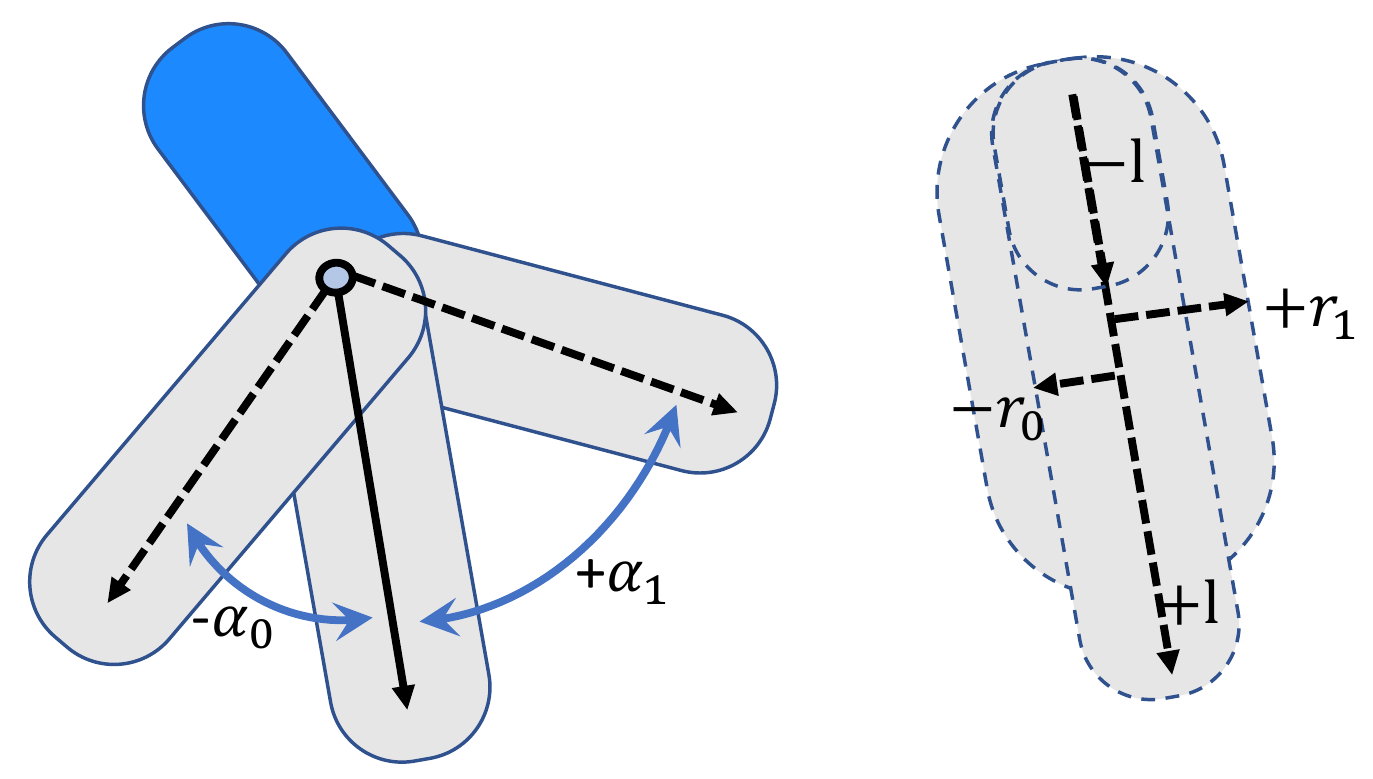}
    \vspace{-3mm}
\end{wrapfigure}

The \textbf{user constraints} (right image) are defined as the ranges of motion, length, radius, etc. Note that the ranges may be asymmetrical. A global constraint defines how much the evolution can change the attributes as a whole. 

\subsection{Physics Simulation and Movement}\label{sec:physics}
The physics of the motion of each agent is simulated with rigid body dynamics. Additionally to the geometric attributes, each edge~$e$ also stores physics attributes: stiffness, damping, friction, and mass density. Each body part also stores its mass, derived from the density and volume. The movement simulation is performed using the Isaac Gym~\cite{makoviychuk2021isaac} which runs parallel physics simulation with multiple environments on GPU. The agent's topology, geometry, and attributes are stored as a \texttt{MJCF} file interpreted by the Isaac Gym. The simulation engine has various parameters. We enable the agent's collision with the environment and self-collision.

The agent's movement is given by the torque $\tau$ applied to each joint over time. There are two methods to control the joint of an agent. The first option (\textit{direct control}) applies the torque directly to each joint, and the actual torque value is provided by the policy network described in the next Sect.~\ref{sec:ppo}. The torque control is fast, but it can be noisy and unstable as the torque is sampled from a policy given distribution. The second option (\textit{PD}) uses Proportional Derivative (PD) controller that works as an intermediate between the control policy and the torque. The control policy specifies the target position for the joint, and the PD provides the torque. This control method provides stable motion as the PD controller can reduce the motion jittering. We use both options in our method and refer to them as \textit{PD} and \textit{direct torque control}.

\subsection{Generation}
We generate the agents either manually or randomly. The manual description is performed by writing the agent description into a text file that is then visualized by the system. The random generation creates the description automatically. It is a two-step process that starts by generating body parts and then attaching legs. The random generation may lead to non-realistic configurations, such as legs inside the body, and they need manual verification for consistency. 
\section{Deep RL Model Representation}\label{sec:ppo}
\begin{figure*}[hbt]
\centering
\includegraphics[width=.99\linewidth]{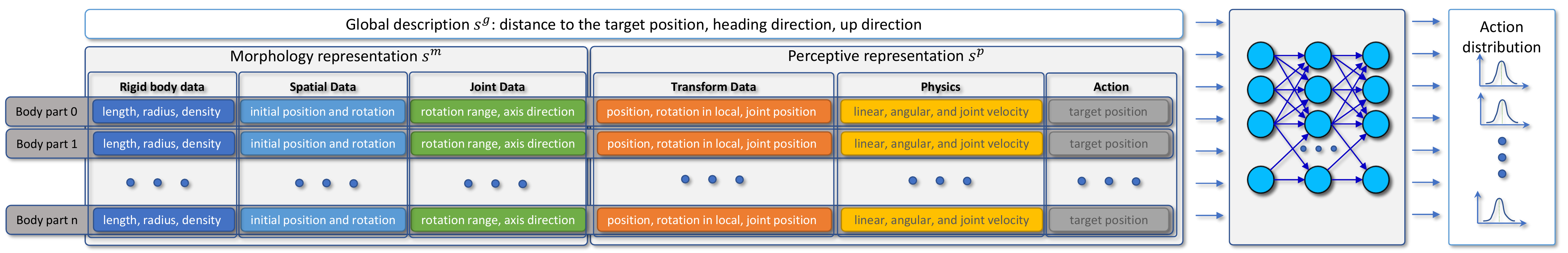}
  \caption{The Control Policy Network (Actor in PPO) of a single agent consisting of multiple body parts. The body part properties and the global description are processed by a deep neural network that generates the corresponding action.}
  \label{fig:agentPPO}\vspace{-3mm}
\end{figure*}
The DeepRL generates a control policy that produces the locomotion for each agent. The learned control policy should be robust across the entire generation. Moreover, we need to train a large number of agents, so the control policy should be able to train rapidly. 

The agent's description and attribute values become the DeepRL framework \textit{states} optimized towards the desired behavior. We use Proximal Policy Optimization (PPO), which is an Actor-Critic-based RL algorithm~\cite{schulman2017proximal}. The Critic estimates the value function and suggests a possible policy distribution, while the Actor updates the policy distribution in the suggested direction. Our universal controller is trained with PPO with advantages computed with Generalized Advantage Estimation~($\lambda$)~\cite{schulman2015high}. The controller receives the state of a agent~$s(t)$ at the time $t$, and it outputs an action $a(t)$ for each joint that leads to the state $s(t+\Delta{t})$. The action $a(t)$ is either the torque $\tau$ applied directly to each joint or a position of a PD controller that then computes the required torque (Sect.~\ref{sec:physics}).

\subsection{States and Actions}
The state of the agent $s(t)$ at time $t$ is (Fig.~\ref{fig:agentPPO}):
\begin{equation}
s(t) = (s^m(t) , s^p(t), s^g(t)),
\end{equation}
where $s^m(t)$ is the agent's morphology, $s^p(t)$ denotes the perceptive representation, and the global representation is denoted by $s^g(t)$. In the following text, we will not specify the time $(t)$, unless needed.

The morphology representation $s^m$ consists of
$$
s^m = \left(s_{rigidbody},s_{spatial},s_{joint}\right),
$$ 
where $s_{rigidbody}$ includes the physics attributes of a body: length, radius, and density. The spatial data $s_{spatial}$ includes the initial direction of the body computed from $fromto$ attributes in \texttt{MJCF} file and the initial local position. The values of $s_{joint}$ contain the attributes of the joints attached to the body, such as the rotation axis and the rotation range of the joint. The morphology representation~$s^m$ does not change during the simulation and training, and it changes only after evolution when the new generation is reproduced (Sect.~\ref{Sec:evo}). Therefore, this part is a constant input to the policy network. The network can then decide on different agents based on their morphology attributes. 

The perceptive representation $s^p$ stores the dynamics information that changes at each time step $t$ 
$$
s^p(t) = (s_{transform},s_{physics},s_{act}),
$$ 
where the transform attributes $s_{transform}$ include the local position, local rotation represented as a quaternion, and the joint position. The physics attributes $s_{physics}$ include linear velocity, angular velocity, and joint velocity. Actions from the previous time step of each joint are also used. The last parameter is the action $s_{act}$ that specifies the target position of the PD controller or direct torques for each joint. The actual value of actions is sampled from Gaussian distributions given by a control policy. We use hinge joints for each agent, specified as the 1D rotation angle $q$ that are normalized based on their joint rotation ranges. Finally, the global description~$s^g$ contains information that indicates the overall behavior of the agent, i.e., distance to the target point, heading direction, and the up vector of the torso.

\subsection{Network Architecture}
The Actor and the Critic in the PPO algorithm are modeled with a deep neural network (see Fig~\ref{fig:agentPPO}). The Actor network is a control policy $\pi$ that maps the given state~$s$ to the Gaussian distributions over actions $\pi(a|s) = \mathcal{N}(\mu(s),\Sigma)$, which takes a mean $\mu(s)$ from the output of the deep neural network and a diagonal covariance matrix $\Sigma$ specified by a vector of learnable parameters~$\vv{\sigma}$. The mean is specified by a fully-connected network with three hidden layers with sizes $[256,128,64]$ and the Exponential Linear Unit (ELU)~\cite{clevert2015fast} as activation function, followed by a linear layer as outputs. The values of covariance matrix $\Sigma= diag(\sigma_0,\sigma_1,...,\sigma_n)$ are learnable parameters and they are updated as a part of the deep neural network with gradient descent. The Critic network~$V(s(t))$ is modeled as a separate network with the same architecture as the Actor network, except that the output size is one, indicating the value of the given state.

\subsection{Rewards}
The reward function $r$ evaluates an agent's performance, e.g., encouraging the agent to walk forward over flat terrain. It attempts to maintain a constant moving speed towards a target distance, and the agent should be able to keep stable locomotion without flipping or deviating from the target direction. It also minimizes energy consumption. The rewards function is a sum of multiple task objectives
\begin{equation}
r = r^p + r^v + r^e + r^a,\label{eqn:rew}
\end{equation}
where $r_p$ is the pose reward that encourages the agent to maintain a stable initial pose during the movement, $r^v$ is the velocity reward, $r^e$~denotes the efficiency reward, and $r^a$ is the alive reward.

The pose reward $r^p$ maintains the heading direction of the agent's body aligned with the target direction $(0,1,0)$ as the agent walks along the $y$-axis. The up direction of the head should point to the up-axis $(0,0,1)$ to prevent the agent swinging its body or flipping:
\begin{equation}
r^p =  w^{heading } \cdot r^{heading}  + w^{up} \cdot r^{up},
\end{equation}
and the weights $w^{heading} = 0.5$ and $w^{up} = 0.1$.
The heading reward~$r^{heading}$ is computed as
\begin{eqnarray}
    p^{heading} & = & \vv{heading} \cdot (0,1,0)\nonumber \\
    r^{heading}& = &\begin{cases}
            1, & \mathrm{if }   p^{heading} \geq t^{heading} \\
            \frac{p^{heading}} { t^{heading}},              & \mathrm{otherwise}
    \end{cases}
\end{eqnarray}
where $p^{heading}$ is the projection of heading vector of the head to the target direction, $t_h = 0.8$ is the threshold of getting the maximum heading reward. We apply the same equation to the up stable reward~$r^{up}$, except that the aligning vector points up and we use a different threshold of $0.9$ that has been established experimentally. 

The velocity reward $r^v$ encourages the agent to move forward along the $y$-axis
\begin{equation}
r^v =  \left(P^y(t)-P^y(t-1)\right)/d_t,
\end{equation}
where $P^y(t)$ is the walking distance along $y$-axis at the time step $t$ and $d_t = 1/60s$.

The efficient reward $r^e$ encourages the agent to perform energy-efficient actions at each time by penalizing high torques or joint close to extreme position to have smoother locomotion. 
\begin{equation}
r^e =  w^{act} \cdot r^{act} +  w^{energy } \cdot r^{energy} + 
            w^{joint limit} \cdot r^{joint limit},
\end{equation}
where the weights are $w^{act } = w^{electricity} = -0.05$ and~$w^{joint limit} = -0.1$. The action cost 
$$
r^{act} = \sum_{\forall joint}{a}^2
$$
penalizes high torque action given by the control policy or joint position closer to the range limitation in the PD control. The energy cost 
$$
r^{energy} = \sum_{\forall joints} |a \cdot v_j|
$$ 
prevents the agent from taking high-energy consumption actions by avoiding high joint velocity $v_j$.

The joint-at-limit reward $r^{jointlimit}$ prevents the agent from not utilizing all joints by penalizing the joint stuck at the limit position
$$
r^{jointlimit} =  w^{jointlimit} \sum_{\forall joint}
\begin{cases}
1, &\text{if } p^{joint} > t^{jointlimit}\\
0, & \text{otherwise}
\end{cases}
$$
where $p_{joint}$ is the normalized joint position, $t^{jointlimit} = 0.99$ is the threshold to receive the penalty and $w^{jointlimit} = -0.1$ is the weight.

The alive reward $r^{a}$ is set to zero when the agent leaves the scene.

\subsection{Training}
Our control policy is trained with proximal-policy optimization (PPO)~\cite{schulman2017proximal} on GPU-based parallel environment Isaac Gym~\cite{makoviychuk2021isaac}. The trained policy is used to evaluate the performance of a variant based on the evaluation method in the previous section.

The training is performed first for the template input agent (Fig.~\ref{fig:overview}~a) and then for each generation during the evolution (Fig.~\ref{fig:overview}~d). Both training stages proceed episodically that start at an initial state~$s_0$ of each agent, which is randomly sampled from a given range to enhance the generalization of the policy network. The experience tuples $(s(t),a(t),r(t),s(t+1))$ are sampled in parallel at each time step $t$ by sampling actions $a$ from control policy $\pi$ with a given state $s(t)$. The experience tuples are stored in the replay buffer for the training iteration later. Each episode is simulated within a maximum number of steps, or it can be terminated by some specific conditions like flipping or wrong walking direction. After the replay buffer is filled with experience tuples, several training iterations are performed to update the Actor network (policy network) and the Critic network (value network). The learning rate~$lr$ is dynamically adapted to the KL-divergence~$kl$ between the new and old policy
\begin{eqnarray}
lr &= &\begin{cases}
\max{(lr/1.5, lr_{min})},       &\mathrm{if\ } kl > \mathrm{desired\ } kl \cdot 2.0\\
\min{(lr \cdot 1.5, lr_{max})}, &\mathrm{if\ } kl > \mathrm{desired\ } kl \cdot 2.0
\end{cases}
\end{eqnarray}
where $lr_{min} = 1e-4$ is the minimum learning rate allowed during the training, $lr_{max} = 1e-3$ is the maximum learning rate, and~$\mathrm{desired\ } kl$ is a hyper-parameter that controls the update of learning rate based on the distance between old policy and the new policy during policy update iteration.

The surrogate loss $L_{surrogate}$ and the value loss $L_{value}$ are clipped within the range defined by the clipping ratio $\epsilon$. 
$$
L_{surrogate/value} = L \cdot clamp (1-\epsilon,1+\epsilon).
$$

\paragraph{Single-agent training}
We train the initial (template) agent (Fig.~\ref{fig:overview}b) to complete the task until the reward Eqn(\ref{eqn:rew}) reaches maximum or does not change significantly. The result provides the baseline policy, the baseline reward value, and the initial locomotion.

\paragraph{Generation Training}
Generation training attempts to optimize a whole generation of agents for evolution. The input to the generation training is the template agent policy. Since each generation of agents shares the same structure, the control policy of the template agent is reused via transfer learning.
 Then, the descendants could quickly inherit the locomotion experience from the previous generation, which in effect, increases the speed of training (to one-fifth in our experiments). 
\begin{figure}[hbt]
\centering
\includegraphics[width=.99\linewidth]{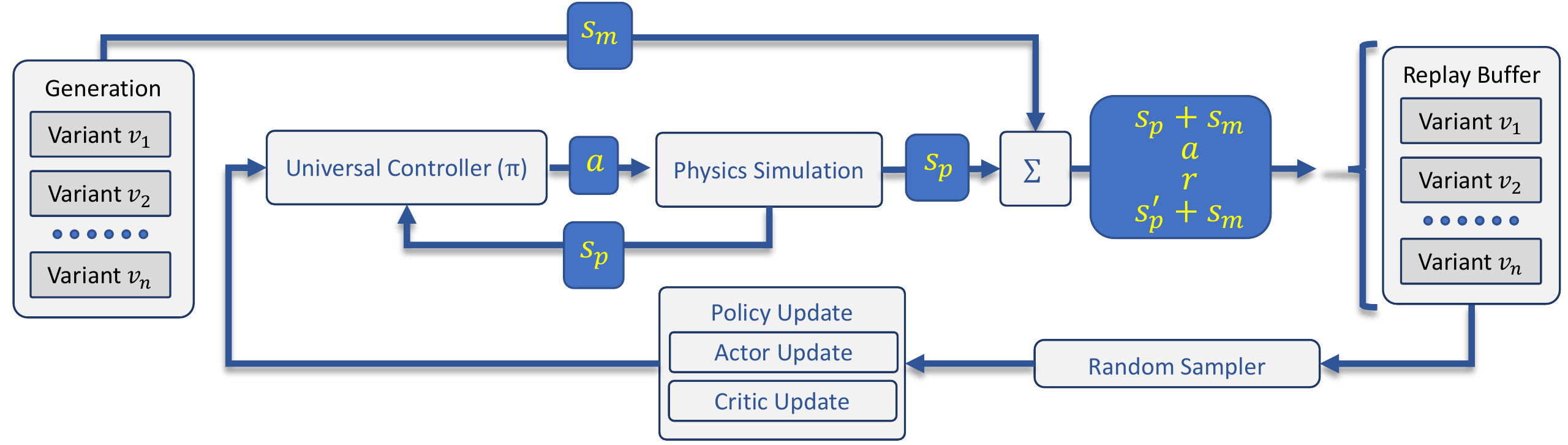}
  \caption{The grouped agent training pipeline where $s_m$ is the morphology state, $s_p$ the receptive state, $a$ are the actions and $r$ rewards.}\label{fig:training}\vspace{-3mm}
\end{figure}

The generation includes $n$ variants trained in parallel (shown in Fig~\ref{fig:training})  each in its environment. At each time step $t$, the universal control policy takes the states $s$ of an agent $v_i$ and outputs its actions~$a$. The experiences are sampled and stored in the replay buffer. The experience tuples sampled from different variants are randomly sampled for the policy update phase. This training part is inspired by metamorph~\cite{gupta2022metamorph} that trains a universal controller with a transformer base structure for robots with different morphology. In our case, we use a simple fully-connected network, providing good performance and training speed.
\section{Evolution}\label{Sec:evo}
Each trained group of agents (Fig.~\ref{fig:training}) produces a set of variants of agents with different body attributes altogether with their reward function. The goal is to choose the best variants of agents and create a new generation while ensuring that their most beneficial traits propagate and possibly improve in the next generation. 

Let $V^g=\{v^g_1,v^g_2,\dots,v^g_n\}$ denote the $g-$th generation with variants of agents~$v_i$. Each agent has a list of attributes $att_i$ that we call its gene. The next generation $g+1$ is produced via selection, crossover, and mutation~\cite{koza95,goldberg2006genetic}. 

\paragraph{Selection:}
We sort all variants $V^g$ in the actual generation $g$ according to their reward and select the top $p\%$ $(p=20)$ agent variants. This initial set becomes the seed of the new generation~$V^{g+1}$.

\paragraph{Crossover:}
The seed of the new generation is expanded to the number of variants $n$ by crossover. We take the genes $att_i$ and $att_j$ of two randomly agent variants $v_i$ and $v_j$ from the seed set. We use a random crossover that takes an attribute $att_i[k]$ and swaps it with~$att_j[k]$ with the 50\% probability. This process is repeated until a new generation $V^{g+1}$ with $n$ variants has been created. 

\paragraph{Mutation:} Each attribute can be mutated by altering its value by a random value $\pm r$. The overall probability of mutation is set to~$1\%$~\cite{goldberg2006genetic}. 

\paragraph{The user-defined constraints:} (Sect.~\ref{sec:shape}) make some attributes fixed, and they will not be affected by the mutation and crossover. Moreover, the values of attributes will not go out of the range of the user-defined constraint limits.

Some attributes can be linked (for example, pair of symmetric legs or body parts belong to the same group (torso body)), and they will always be treated as a fixed group. When one of them is swapped, the other will be as well. If one value is changed, the others will be changed by the same value.
\section{Implementation and Results}

\subsection{Implementation}
We use Python to develop the agent generator and all the components in our evolution system. Isaac Gym~\cite{makoviychuk2021isaac} was used for the physics simulation of the robot, and we implemented the PPO optimization in Python. The neural network is based on Pytorch version 1.8.1. The computation, including deep neural network training, physics simulation, and rendering, runs on a single Nvidia GeForce RTX 3090. The baseline agent is trained for 500 epochs with 900 parallel environments, and the entire training takes approximately 10 minutes. The agent generation training with the universal controller is trained for 35 epochs and 150 variants. Each variant runs on six parallel environments. The training for each generation takes around 60 seconds. The overall evolution of the 50 generations takes around 40 minutes to 60 minutes, depending on the complexity of the agent and the environment. The main limitation is the size of the GPU memory.

\subsection{Results}
We designed and randomly generated several agents to test the effect of the evolution on the agent co-design. All results are summarized in Tab.~\ref{table:results}, and details of each body part are in the Appendix. Please note this paper has the accompanying video that shows the results from this paper.

\begin{table}[htb]
\scriptsize
\begin{tabular}{lclrcc}
\toprule
& \multicolumn{2}{c}{\textbf{Constrained}} & \textbf{Reward} & \textbf{New Reward (\%)} & \textbf{Actual Change (\%)} \\
 & \textbf{Fix Body}         & \textbf{Evolution}        &        &                  &                    \\
\midrule
Fig. \ref{fig:teaser} (a) & - & 20\% & 960 & 100\% (baseline)&   N/A  \\
Fig. \ref{fig:teaser} (b) & No & 20\% & 1,274 & 132\% & 0.39\%    \\
Fig. \ref{fig:teaser} (c) & No & 20\% & 1,594 & 166\% &  5.82\%   \\
Fig. \ref{fig:teaser} (d) & No & 20\% & 1,775 & 184\% &  5.84\%   \\
Fig. \ref{fig:teaser} (e) & No & 20\% & 1,913 & 199\% &  5.63\%   \\
Fig. \ref{fig:teaser} (f) & No & 20\% & 2,169 & 228\% & 5.72\%    \\
Fig. \ref{fig:teaser} (g) & No & 20\% & 1,160 & 121\% &  0.56\%   \\
Fig. \ref{fig:teaser} (h) & No & 20\% & 1,842 & 192\% &  8.93\%   \\
Fig. \ref{fig:teaser} (i) & No & 20\% & 2,174 & 226\% & 9.40\%    \\
Fig. \ref{fig:teaser} (j) & No & 20\% & 2,355 & 245\% &  9.45\%   \\
Fig. \ref{fig:teaser} (k) & No & 20\% & 2,428 & 253\% &  9.71\%   \\
Fig. \ref{fig:expA} (a) & - & 0\% & 470 & 100\% (baseline) &   N/A  \\
Fig. \ref{fig:expA} (b) & No & 10\% & 621 & 132\% & 3.24\%    \\
Fig. \ref{fig:expA} (c) & No & 20\% & 710 & 151\% & 8.75\%    \\
Fig. \ref{fig:expBa} (base) & - & 0\% & 489 & 100\% (baseline) &   N/A   \\
Fig. \ref{fig:expBa} (evo) & No & 20\% & 566 & 116\% & 10.83\%    \\
Fig. \ref{fig:expBb} (base) & - & 0\% & 572 & 100\% (baseline) &  N/A   \\
Fig. \ref{fig:expBb} (evo) & No & 20\% & 921 & 161\% & 8.02\%    \\
Fig. \ref{fig:expBc} (base) & - & 0\% & 683 & 100\%  (baseline)&  N/A   \\
Fig. \ref{fig:expBc} (evo) &No  & 20\% & 1,108 & 155\% & 2.47\%    \\
Fig. \ref{fig:expD} (a) & - & 0\% & 683 & 100\%  (baseline)&  N/A   \\
Fig. \ref{fig:expD} (b) & torso & 40\% & 1,108 & 162\% & 5.24\%    \\
Fig. \ref{fig:expD} (c) & leg & 40\% & 870 & 127 & 6.44\%    \\ \bottomrule
\end{tabular}
\caption{Quantitative results of all experiments.}\label{table:results}\vspace{-3mm}
\end{table}

The first example in Fig.~\ref{fig:teaser} shows the effect of the evolution on the changes and reward function of an agent. The baseline agent is trained to walk with the state-of-the-art PPO training (a), and we then use the evolutionary algorithm to improve its performance while changing its attributes to complete the same task. The reward function value for the baseline agent is 473, and it improves through the evolution after the first generation to 132\% (b), the fifth generation 166\% (c), 15-th generation 184\% (d), 25-th generation 199\% (e), and 35-th generation to 228\% (f). We then take the same agent and fix its body shape so it cannot change through evolution. The agent is trained from the baseline leading to the new reward after the first generation to 121\% (g), the fifth generation 192\% (h), 15-th generation 226\% (i), 25-th generation 245\% (j), and 35-th generation to 253\% (k).

\begin{figure}[hbt]
\centering
\includegraphics[width=0.99\linewidth]{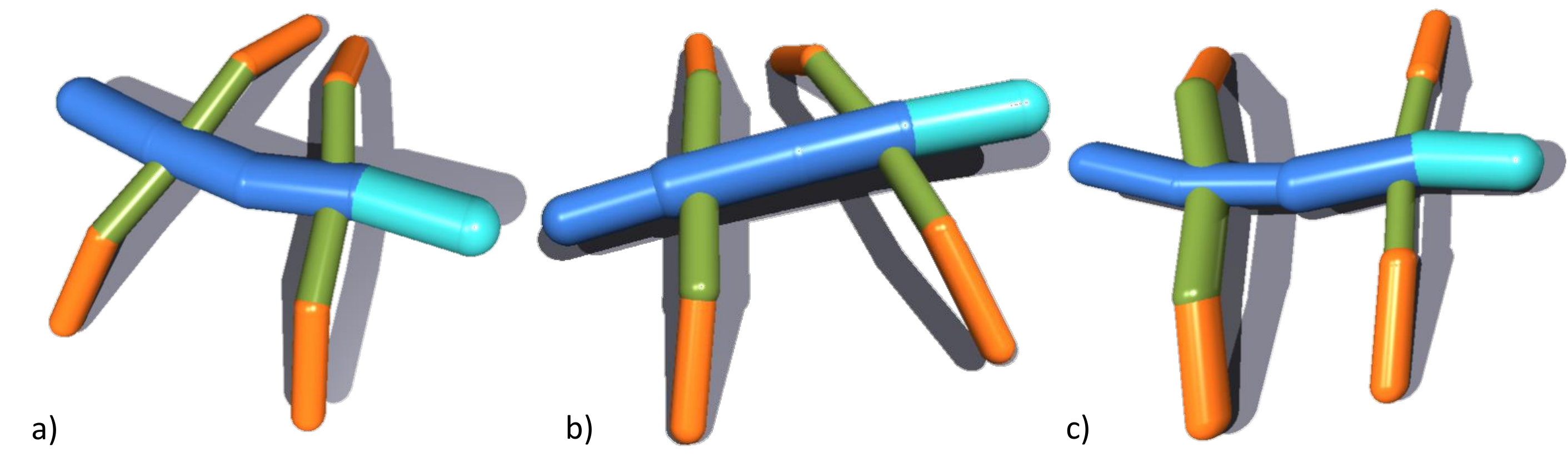}
\caption{A baseline agent (a) is evolved by allowing $\pm10\%$ (b), and $\pm20\%$ (c) of variance of all its parameters. The reward function value 470 of the baseline agent (a) improves to 132\% (b), and 151\% (c).}\label{fig:expA}\vspace{-3mm}
\end{figure}

The experiment in Fig.~\ref{fig:expA} studies the effect of globally increasing the range of allowed changes. The baseline input agent has been trained, leading to the reward function value of 470. We then run the evolutionary co-design, allowing the global change attributes by $\pm10\%$ and $\pm20\%$. While the reward is increasing to 132, and 151\% of the baseline design, the structure of the agent has also changed significantly. 

\begin{figure}[hbt]
\centering
\includegraphics[width=0.99\linewidth]{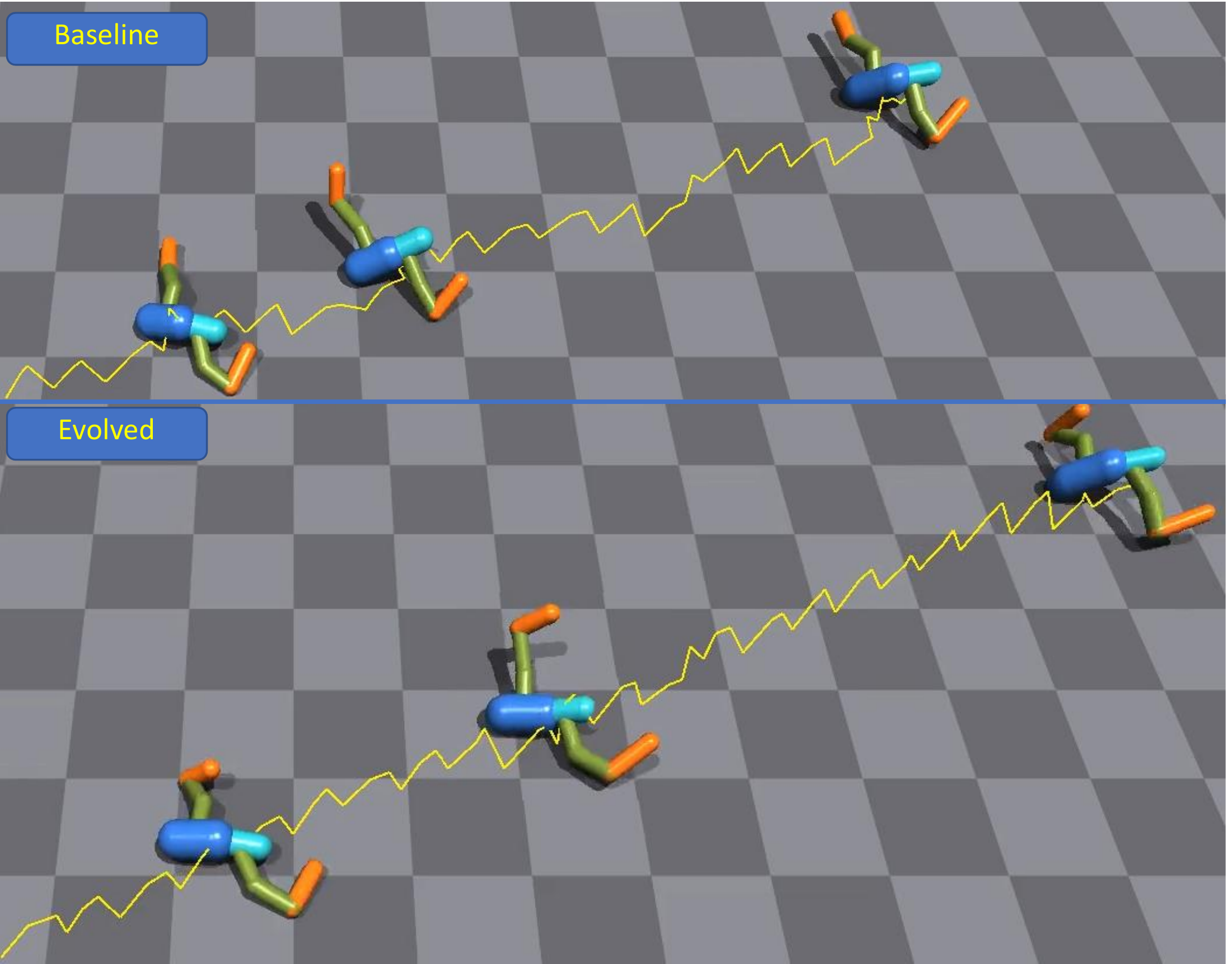}
\caption{A simple baseline agent (top) evolved by allowing $\pm 20\%$ of variance of all its parameters. The evolved agents travel larger distance in the same time and the evolved reward functions are improved  489$\rightarrow$566 (116\%).}\label{fig:expBa}\vspace{-3mm}
\end{figure}

\begin{figure}[hbt]
\centering
\includegraphics[width=0.99\linewidth]{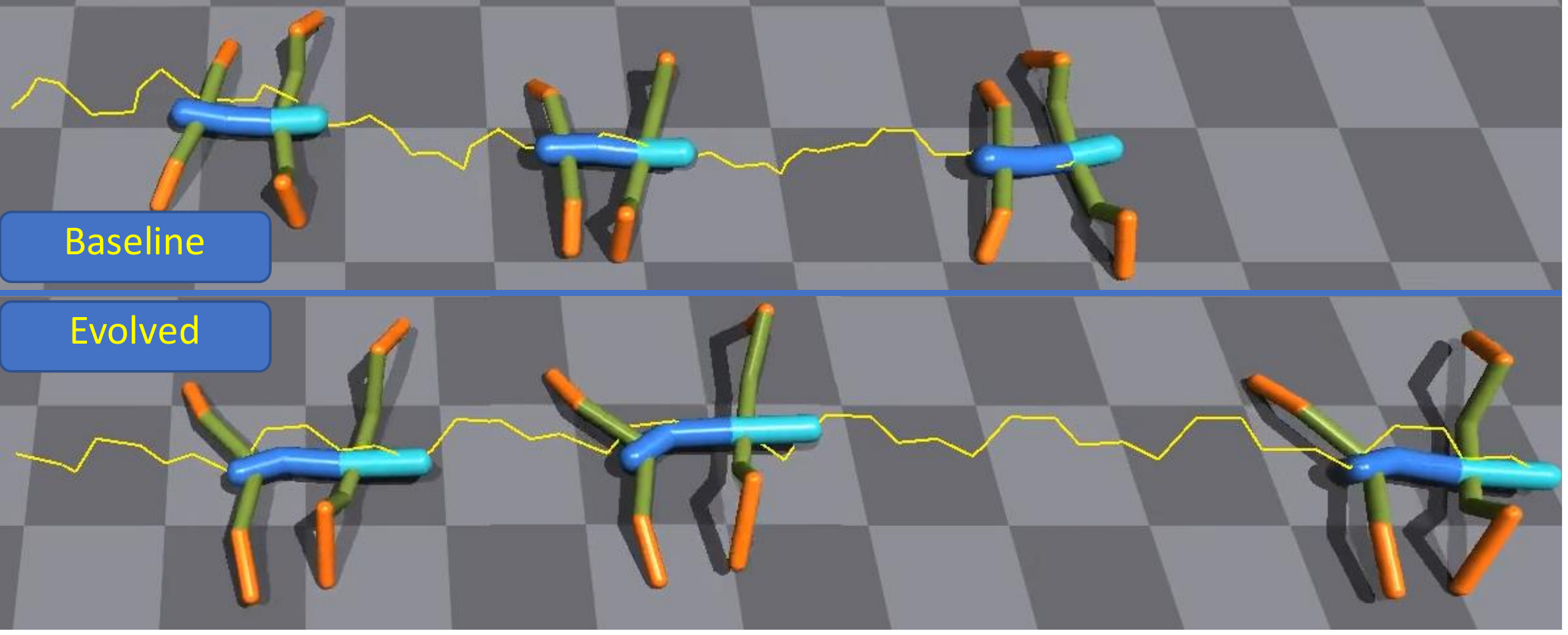}
\caption{A medium complex baseline agent (top) evolved by allowing $\pm 20\%$ of variance of its parameters. The evolved reward functions are 572$\rightarrow$921 (161\%).}\label{fig:expBb}\vspace{-3mm}
\end{figure}

\begin{figure}[hbt]
\centering
\includegraphics[width=0.99\linewidth]{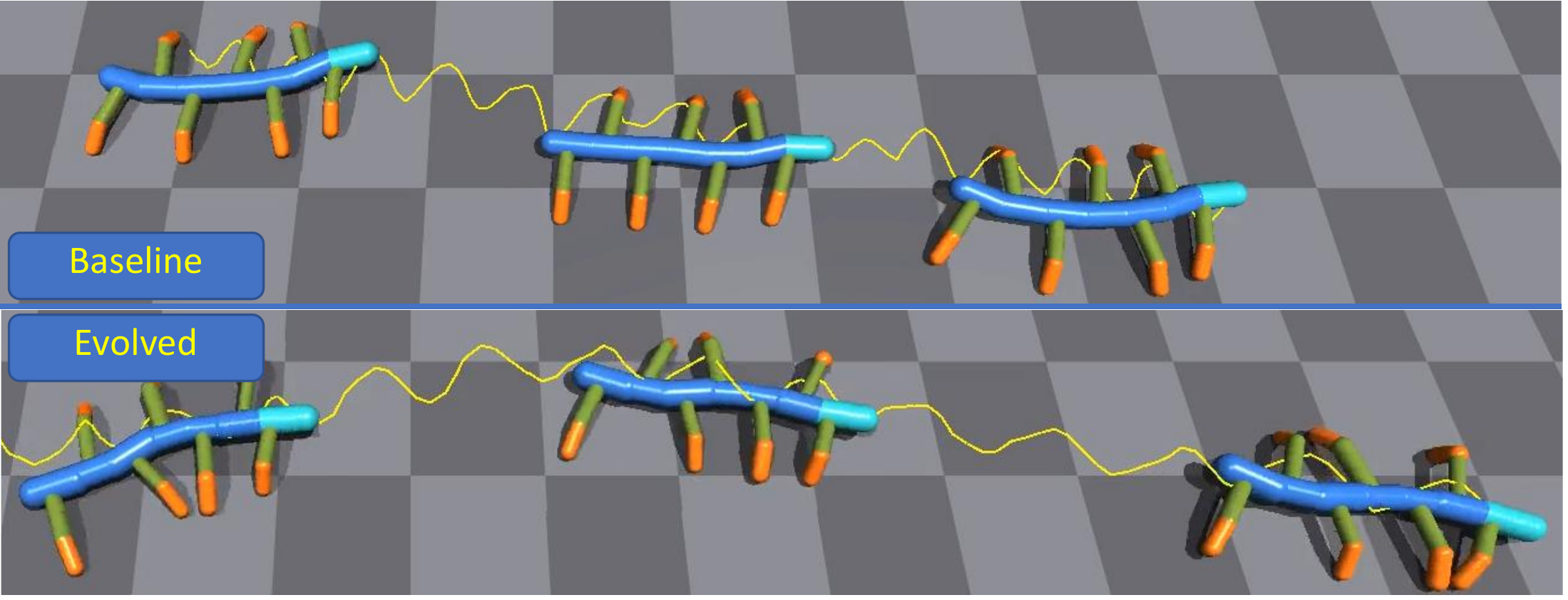}
\caption{A complex baseline agent (top) evolved by allowing $\pm 20\%$ of variance. The evolved reward functions are 1,118$\rightarrow$1,737 (155\%).}\label{fig:expBc}\vspace{-3mm}
\end{figure}

Figures~\ref{fig:expBa}-\ref{fig:expBc} show three agents with increasing complexity evolved by allowing $\pm 20\%$ of global attributes changes. The snapshots of the motion are taken after the same time, showing the traveled distance for comparison. The simple agent improved to 153\% of the baseline model, the medium to 161\%, and the complex one to~155\%.  

\begin{figure}[hbt]
\centering
\includegraphics[width=0.99\linewidth]{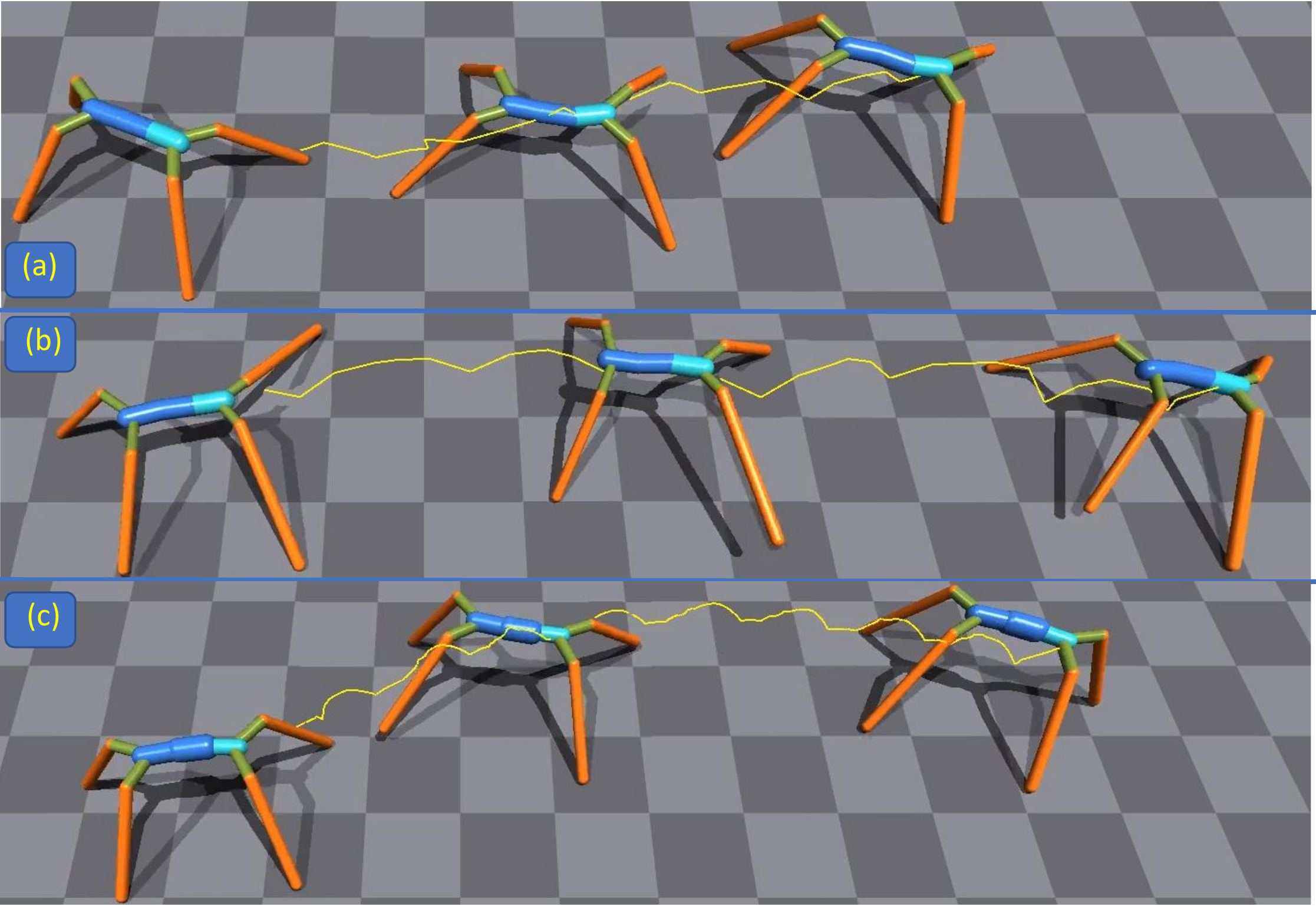}
\caption{An agent (a) is evolved with restricted torso and allowed changes of 40\% to the rest of the body. The legs improved, and the reward function changed 683$\rightarrow$1,108 (162\%) (b). The last row shows the same creature evolved only with allowed modifications to the torso (legs are fixed) reward function changed 683$\rightarrow$870 (127\%) (c)}\label{fig:expD}\vspace{-3mm}
\end{figure}

Another example in Fig.~\ref{fig:expD} shows the effect of the restricted control of the evolution. We fixed the torso (Fig.~\ref{fig:expD}~a) during the evolution by not allowing any changes in the agent. While the body remains the same, the legs and their control were allowed to change by 40\%, leading to the improvement of 162\%. Fig.~\ref{fig:expD}~b) shows the same agent where only the torso can evolve, and the legs remain fixed. This limits the motion, and the improvement was only 127\% of the baseline.

While the above-mentioned examples were generated with the PD control, the accompanying video shows that our evolutionary algorithm handles the direct torque control from the PPO. 

\begin{figure}[hbt]
\centering
\includegraphics[width=0.99\linewidth]{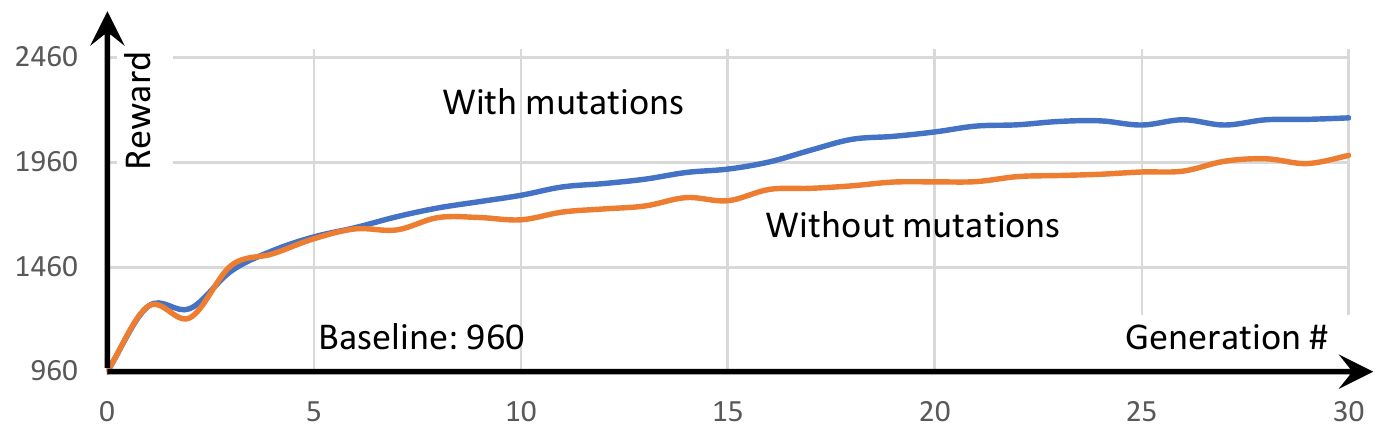}
\caption{The agent from Fig.~\ref{fig:teaser} is evolved with and without the mutation showing that the mutations have a positive effect on the reward function.}\label{fig:mutation}\vspace{-3mm}
\end{figure}

We tested the effect of the mutation on the convergence of the reward function. We trained the baseline agent from Fig.~\ref{fig:teaser} with and without the mutation. The progress of both reward functions in Fig.~\ref{fig:mutation} shows that the mutation has a positive effect on the reward function leading to faster convergence and about 9\%  higher reward (2,171	vs. 1992).

\begin{figure}[hbt]
\centering
\includegraphics[width=0.99\linewidth]{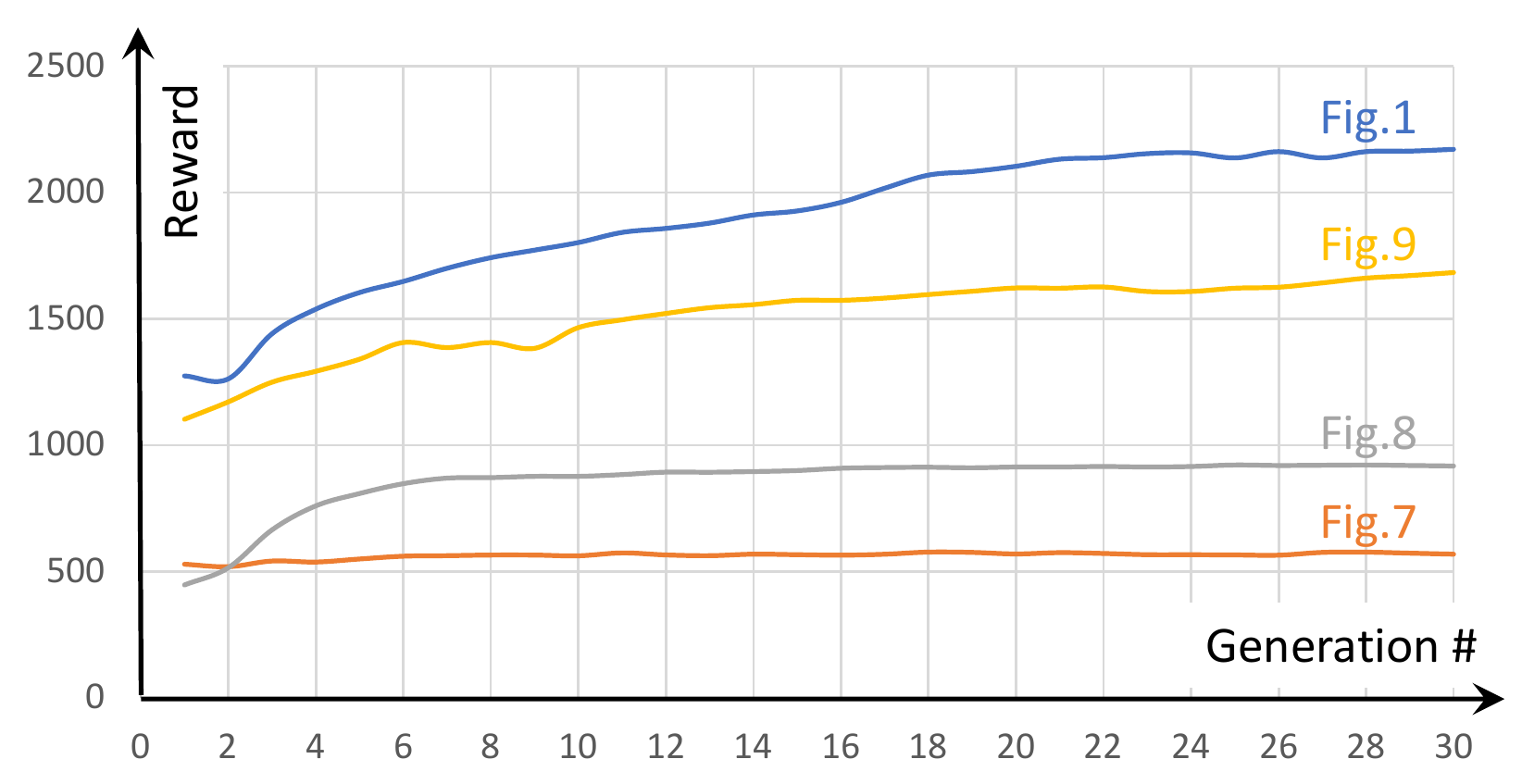}
\caption{Reward function evolution for the examples from this paper.}\label{fig:allRewards}\vspace{-3mm}
\end{figure}

The reward functions through the 30 generations of the evolution for figures in this paper are shown in Fig.~\ref{fig:allRewards}. The reward function increases most if no constraints are imposed on the model, or if the model has high complexity allowing for more changes. 

We attempted to provide some insight into the traits that affect the overall performance of the agents. We analyzed the data from the Appendix that show all number of changes for agents from Figs.~\ref{fig:expBa}, \ref{fig:expBb}, \ref{fig:expBc}, and~\ref{fig:expD}. The overall tendency allowing the agents to perform better is diminishing their weight. The control parameters play an important role in the locomotion as its global changes are relatively higher than the others. The statistics show that the increase in the body's average length also helps improve performance. This is especially true for the legs, indicating that longer legs are beneficial. Moreover, stiffness and the max effort tend to increase through the evolution as they provide a faster response to the target joint position, and they increase the maximum torque. An exception is an agent in Fig.~\ref{fig:expD} that could not evolve its legs, leading to decreased damping and the max effort. 

\section{Conclusions, Limitations, and Future Work}
We have introduced a novel approach that improves the state-of-the-art DeepRL training by adding evolutionary changes to the agent's parameters. While the agent's topology remains the same, the genetic algorithm explores the space of the agent's attributes and attempts to improve its performance to complete the given task. Our approach has two main advantages. First, it allows for user control of the evolving parts. Second, it uses a universal policy and transfer learning that enables us to train a whole generation of agents on a single GPU. This significantly shortens the training time of the evolutionary algorithm to 1 minute per generation. We have shown various examples of agents trained with varying shapes and parameters, showing that the performance improved by tens of percent even after just a few generations. 

Our approach has several limitations. First, we used Isaac Gym and PPO as our simulation and RL training baseline. While this is a fair choice, both RL algorithms and physics engine include many parameters that need to be carefully tuned, and they may have a negative effect on the training. We have carefully used precisely the same parameters when comparing the results, but we noted, for example, that using self-collision detection for complex agents changes the results significantly. The second limitation is the improvement of evolutionary requires the template agent is able to perform the task to provide an initial control policy. If the template agent fails the task, the descendants will not benefit from the pre-trained policy.

There are many possible avenues for future work. First, it would be interesting to study how many and what parameters are suitable for the user. We showed several ways of controlling the shape and its evolution, but the actual user intent and feedback would be worthy of its research project. Second, the space that needs to be explored during the evolution is vast, and it is evident that our approach is leading only to a limited set of solutions. Future work could use several solutions and see what makes them different. Another important problem to study in the future is to answer the question of what makes the design perform better. It could be achieved by tracking the values of attributes and seeing how they relate to the performance. However, the relation is very unlikely straightforward, and the parameters may affect each other. Obvious future work is studying more complex tasks and environments. 

\bibliographystyle{ACM-Reference-Format}
\bibliography{main}

\newpage
\appendix
\section{Appendix}

\setcounter{table}{0}
\renewcommand{\thetable}{A\arabic{table}}

\begin{table*}[htb]
\begin{tabular}{lllllll}
\toprule
 &	\textbf{Length} &	\textbf{Radius}	&\textbf{Density} & \textbf{Stiffness} & \textbf{Damping} & \textbf{Max Effort}\\
\midrule
\multicolumn{7}{c}{\textbf{Baseline}}                                                                                                   \\
\midrule
\textbf{avg}   & \textbf{0.40} & \textbf{0.07} & \textbf{5.00} & \textbf{5.00}  & \textbf{2.00}  & \textbf{5.00}                        \\
\midrule
body 0         & 0.40          & 0.10          & 5.00          & 5.00           & 2.00           & 5.00                                 \\
body 1         & 0.40          & 0.10          & 5.00          & 5.00           & 2.00           & 5.00                                 \\
body 2         & 0.40          & 0.06          & 5.00          & 5.00           & 2.00           & 5.00                                 \\
body 3         & 0.40          & 0.06          & 5.00          & 5.00           & 2.00           & 5.00                                 \\
body 4         & 0.40          & 0.06          & 5.00          & 5.00           & 2.00           & 5.00                                 \\
body 5         & 0.40          & 0.06          & 5.00          & 5.00           & 2.00           & 5.00                                 \\
body 6         & 0.40          & 0.10          & 5.00          & 5.00           & 2.00           & 5.00                                 \\
body 7         & 0.40          & 0.06          & 5.00          & 5.00           & 2.00           & 5.00                                 \\
body 8         & 0.40          & 0.06          & 5.00          & 5.00           & 2.00           & 5.00                                 \\
body 9         & 0.40          & 0.06          & 5.00          & 5.00           & 2.00           & 5.00                                 \\
body 10        & 0.40          & 0.06          & 5.00          & 5.00           & 2.00           & 5.00                                 \\
body 11        & 0.40          & 0.10          & 5.00          & 5.00           & 2.00           & 5.00                                 \\
\midrule
\multicolumn{7}{c}{\textbf{Evolved} $\pm10\%$ (Agent in Fig.  \ref{fig:expA} (b))}                                                                                             \\
\midrule
\textbf{avg}   & \textbf{0.41} & \textbf{0.07} & \textbf{4.90} & \textbf{5.27}  & \textbf{1.90}  & \textbf{4.85}                        \\
\midrule
body 0         & 0.42          & 0.10          & 5.35          & 5.27           & 1.90           & 4.85                                 \\
body 1         & 0.38          & 0.11          & 4.85          & 5.27           & 1.90           & 4.85                                 \\
body 2         & 0.41          & 0.06          & 5.16          & 5.27           & 1.90           & 4.85                                 \\
body 3         & 0.41          & 0.06          & 5.16          & 5.27           & 1.90           & 4.85                                 \\
body 4         & 0.42          & 0.06          & 4.99          & 5.27           & 1.90           & 4.85                                 \\
body 5         & 0.42          & 0.06          & 4.99          & 5.27           & 1.90           & 4.85                                 \\
body 6         & 0.43          & 0.11          & 4.51          & 5.27           & 1.90           & 4.85                                 \\
body 7         & 0.41          & 0.07          & 4.32          & 5.27           & 1.90           & 4.85                                 \\
body 8         & 0.41          & 0.07          & 4.32          & 5.27           & 1.90           & 4.85                                 \\
body 9         & 0.41          & 0.06          & 5.19          & 5.27           & 1.90           & 4.85                                 \\
body 10        & 0.41          & 0.06          & 5.19          & 5.27           & 1.90           & 4.85                                 \\
body 11        & 0.38          & 0.09          & 4.78          & 5.27           & 1.90           & 4.85                                 \\
\midrule
\textbf{Delta} & \textbf{2.18} & \textbf{1.79} & \textbf{1.99} & \textbf{5.09}  & \textbf{5.26}  & \textbf{3.12}                        \\
\midrule
\multicolumn{6}{r}{\textbf{Actual Change:}}                                                       & {\color[HTML]{FF0000} \textbf{3.24\%}} \\
\midrule
\multicolumn{7}{c}{\textbf{Evolved} $\pm20\%$ (Agent in Fig.  \ref{fig:expA} (c))}                                                                                             \\
\midrule
\textbf{avg}   & \textbf{0.43} & \textbf{0.08} & \textbf{4.91} & \textbf{5.93}  & \textbf{2.24}  & \textbf{5.97}                        \\
\midrule
body 0         & 0.36          & 0.12          & 5.77          & 5.93           & 2.24           & 5.97                                 \\
body 1         & 0.46          & 0.11          & 5.45          & 5.93           & 2.24           & 5.97                                 \\
body 2         & 0.41          & 0.05          & 5.29          & 5.93           & 2.24           & 5.97                                 \\
body 3         & 0.41          & 0.05          & 5.29          & 5.93           & 2.24           & 5.97                                 \\
body 4         & 0.42          & 0.07          & 4.76          & 5.93           & 2.24           & 5.97                                 \\
body 5         & 0.42          & 0.07          & 4.76          & 5.93           & 2.24           & 5.97                                 \\
body 6         & 0.42          & 0.08          & 4.50          & 5.93           & 2.24           & 5.97                                 \\
body 7         & 0.42          & 0.07          & 3.51          & 5.93           & 2.24           & 5.97                                 \\
body 8         & 0.42          & 0.07          & 3.51          & 5.93           & 2.24           & 5.97                                 \\
body 9         & 0.53          & 0.07          & 5.55          & 5.93           & 2.24           & 5.97                                 \\
body 10        & 0.53          & 0.07          & 5.55          & 5.93           & 2.24           & 5.97                                 \\
body 11        & 0.35          & 0.08          & 5.00          & 5.93           & 2.24           & 5.97                                 \\
\midrule
\textbf{Delta} & \textbf{4.37} & \textbf{2.69} & \textbf{0.19} & \textbf{11.22} & \textbf{15.30} & \textbf{18.74}                       \\
\midrule
\multicolumn{6}{r}{\textbf{Actual Change:}}                                                       & {\color[HTML]{FF0000} \textbf{8.75\%}}\\
\bottomrule
\end{tabular}
\caption{Detailed comparison between baseline (Fig.  \ref{fig:expA} (a)) and evolved agents (Fig.  \ref{fig:expA} (b) and Fig.  \ref{fig:expA} (c)).}\label{table:expA}
\end{table*}

\begin{table*}[htb]
\begin{tabular}{lllllll}
\toprule
 &	\textbf{Length} &	\textbf{Radius}	&\textbf{Density} & \textbf{Stiffness} & \textbf{Damping} & \textbf{Max Effort}\\
\midrule
\multicolumn{7}{c}{\textbf{Baseline}}                                                                                                   \\
\midrule
\textbf{avg}   & \textbf{0.34} & \textbf{0.09}  & \textbf{5.17}  & \textbf{5.00} & \textbf{2.00}           & \textbf{10.00}                        \\
\midrule
body 0         & 0.33          & 0.12           & 5.85           & 5.00          & 2.00                    & 10.00                                 \\
body 1         & 0.32          & 0.17           & 6.68           & 5.00          & 2.00                    & 10.00                                 \\
body 2         & 0.38          & 0.08           & 3.10           & 5.00          & 2.00                    & 10.00                                 \\
body 3         & 0.31          & 0.08           & 5.74           & 5.00          & 2.00                    & 10.00                                 \\
body 4         & 0.38          & 0.08           & 6.02           & 5.00          & 2.00                    & 10.00                                 \\
body 5         & 0.33          & 0.08           & 3.27           & 5.00          & 2.00                    & 10.00                                 \\
body 6         & 0.40          & 0.08           & 3.79           & 5.00          & 2.00                    & 10.00                                 \\
body 7         & 0.30          & 0.08           & 6.87           & 5.00          & 2.00                    & 10.00                                 \\
\midrule
\multicolumn{7}{c}{\textbf{Evolved} $\pm20\%$}                                                                                             \\
\midrule
\textbf{avg}   & \textbf{0.36} & \textbf{0.09}  & \textbf{5.11}  & \textbf{5.12} & \textbf{2.09}           & \textbf{10.45}                        \\
\midrule
body 0         & 0.36          & 0.11           & 5.80           & 5.12          & 2.09                    & 10.45                                 \\
body 1         & 0.33          & 0.17           & 6.15           & 5.12          & 2.09                    & 10.45                                 \\
body 2         & 0.40          & 0.07           & 3.39           & 5.12          & 2.09                    & 10.45                                 \\
body 3         & 0.34          & 0.08           & 5.64           & 5.12          & 2.09                    & 10.45                                 \\
body 4         & 0.39          & 0.08           & 5.84           & 5.12          & 2.09                    & 10.45                                 \\
body 5         & 0.35          & 0.08           & 3.42           & 5.12          & 2.09                    & 10.45                                 \\
body 6         & 0.42          & 0.07           & 3.47           & 5.12          & 2.09                    & 10.45                                 \\
body 7         & 0.29          & 0.08           & 7.18           & 5.12          & 2.09                    & 10.45                                 \\
\midrule
\textbf{Delta} & \textbf{7.90} & \textbf{31.63} & \textbf{14.53} & \textbf{2.40} & \textbf{4.20}           & \textbf{4.32}                         \\
\midrule
\multicolumn{6}{r}{\textbf{Actual Change:}}                                                       & {\color[HTML]{FF0000} \textbf{10.83\%}} \\
\bottomrule
\end{tabular}
\caption{Detailed comparison between the baseline and evolved agent of Fig.  \ref{fig:expBa}.}\label{table:expBa}
\end{table*}

\begin{table*}[htb]
\begin{tabular}{lllllll}
\toprule
 &	\textbf{Length} &	\textbf{Radius}	&\textbf{Density} & \textbf{Stiffness} & \textbf{Damping} & \textbf{Max Effort}\\
\midrule
\multicolumn{7}{c}{\textbf{Baseline}}                                                                                                   \\
\midrule
\textbf{avg}   & \textbf{0.35}  & \textbf{0.07} & \textbf{5.00} & \textbf{5.00} & \textbf{2.00}  & \textbf{5.00} \\
\midrule
body 0         & 0.30           & 0.10          & 5.00          & 5.00          & 2.00           & 5.00          \\
body 1         & 0.40           & 0.06          & 5.00          & 5.00          & 2.00           & 5.00          \\
body 2         & 0.40           & 0.06          & 5.00          & 5.00          & 2.00           & 5.00          \\
body 3         & 0.30           & 0.06          & 5.00          & 5.00          & 2.00           & 5.00          \\
body 4         & 0.30           & 0.06          & 5.00          & 5.00          & 2.00           & 5.00          \\
body 5         & 0.40           & 0.06          & 5.00          & 5.00          & 2.00           & 5.00          \\
body 6         & 0.40           & 0.06          & 5.00          & 5.00          & 2.00           & 5.00          \\
body 7         & 0.30           & 0.10          & 5.00          & 5.00          & 2.00           & 5.00          \\
body 8         & 0.30           & 0.10          & 5.00          & 5.00          & 2.00           & 5.00          \\
body 9         & 0.40           & 0.06          & 5.00          & 5.00          & 2.00           & 5.00          \\
body 10        & 0.40           & 0.06          & 5.00          & 5.00          & 2.00           & 5.00          \\
body 11        & 0.30           & 0.06          & 5.00          & 5.00          & 2.00           & 5.00          \\
body 12        & 0.30           & 0.06          & 5.00          & 5.00          & 2.00           & 5.00                                 \\
\midrule
\multicolumn{7}{c}{\textbf{Evolved} $\pm20\%$}                                                                                             \\
\midrule
\textbf{avg}   & \textbf{0.47}  & \textbf{0.07} & \textbf{4.92} & \textbf{4.84} & \textbf{2.19}  & \textbf{5.47} \\
\midrule
body 0         & 0.56           & 0.10          & 4.34          & 4.84          & 2.19           & 5.47          \\
body 1         & 0.34           & 0.06          & 4.64          & 4.84          & 2.19           & 5.47          \\
body 2         & 0.47           & 0.06          & 3.01          & 4.84          & 2.19           & 5.47          \\
body 3         & 0.47           & 0.06          & 5.67          & 4.84          & 2.19           & 5.47          \\
body 4         & 0.47           & 0.06          & 3.19          & 4.84          & 2.19           & 5.47          \\
body 5         & 0.59           & 0.06          & 6.88          & 4.84          & 2.19           & 5.47          \\
body 6         & 0.59           & 0.06          & 6.88          & 4.84          & 2.19           & 5.47          \\
body 7         & 0.42           & 0.10          & 2.59          & 4.84          & 2.19           & 5.47          \\
body 8         & 0.31           & 0.10          & 4.54          & 4.84          & 2.19           & 5.47          \\
body 9         & 0.52           & 0.06          & 4.18          & 4.84          & 2.19           & 5.47          \\
body 10        & 0.52           & 0.06          & 5.98          & 4.84          & 2.19           & 5.47          \\
body 11        & 0.46           & 0.06          & 6.06          & 4.84          & 2.19           & 5.47          \\
body 12        & 0.37           & 0.06          & 6.06          & 4.84          & 2.19           & 5.47                                 \\
\midrule
\textbf{Delta} & \textbf{26.10} & \textbf{0.00} & \textbf{1.53} & \textbf{3.39} & \textbf{8.48}  & \textbf{8.61} \\
\midrule
\multicolumn{6}{r}{\textbf{Actual Change:}}                                                       & {\color[HTML]{FF0000} \textbf{8.02\%}} \\
\bottomrule
\end{tabular}
\caption{Detailed comparison between the baseline and evolved agent of Fig.  \ref{fig:expBb}.}\label{table:expBb}
\end{table*}

\begin{table*}[htb]
\begin{tabular}{lllllll}
\toprule
 &	\textbf{Length} &	\textbf{Radius}	&\textbf{Density} & \textbf{Stiffness} & \textbf{Damping} & \textbf{Max Effort}\\
\midrule
\multicolumn{7}{c}{\textbf{Baseline}}                                                                                                   \\
\midrule
\textbf{avg}   & \textbf{0.30}       & \textbf{0.07}       & \textbf{5.00}       & \textbf{50.00}      & \textbf{15.00}          & \textbf{500.00}     \\
\midrule
body 0         & 0.30                & 0.10                & 5.00                & 50.00               & 15.00                   & 500.00              \\
body 1         & 0.40                & 0.06                & 5.00                & 50.00               & 15.00                   & 500.00              \\
body 2         & 0.20                & 0.06                & 5.00                & 50.00               & 15.00                   & 500.00              \\
body 3         & 0.30                & 0.10                & 5.00                & 50.00               & 15.00                   & 500.00              \\
body 4         & 0.40                & 0.06                & 5.00                & 50.00               & 15.00                   & 500.00              \\
body 5         & 0.20                & 0.06                & 5.00                & 50.00               & 15.00                   & 500.00              \\
body 6         & 0.30                & 0.10                & 5.00                & 50.00               & 15.00                   & 500.00              \\
body 7         & 0.40                & 0.06                & 5.00                & 50.00               & 15.00                   & 500.00              \\
body 8         & 0.20                & 0.06                & 5.00                & 50.00               & 15.00                   & 500.00              \\
body 9         & 0.30                & 0.10                & 5.00                & 50.00               & 15.00                   & 500.00              \\
body 10        & 0.40                & 0.06                & 5.00                & 50.00               & 15.00                   & 500.00              \\
body 11        & 0.20                & 0.06                & 5.00                & 50.00               & 15.00                   & 500.00              \\
body 12        & 0.30                & 0.10                & 5.00                & 50.00               & 15.00                   & 500.00              \\
body 13        & 0.40                & 0.06                & 5.00                & 50.00               & 15.00                   & 500.00              \\
body 14        & 0.20                & 0.06                & 5.00                & 50.00               & 15.00                   & 500.00              \\
body 15        & 0.30                & 0.10                & 5.00                & 50.00               & 15.00                   & 500.00              \\
body 16        & 0.40                & 0.06                & 5.00                & 50.00               & 15.00                   & 500.00              \\
body 17        & 0.20                & 0.06                & 5.00                & 50.00               & 15.00                   & 500.00              \\
body 18        & 0.30                & 0.10                & 5.00                & 50.00               & 15.00                   & 500.00              \\
body 19        & 0.40                & 0.06                & 5.00                & 50.00               & 15.00                   & 500.00              \\
body 20        & 0.20                & 0.06                & 5.00                & 50.00               & 15.00                   & 500.00                                 \\
\midrule
\multicolumn{7}{c}{\textbf{Evolved} $\pm20\%$}                                                                                             \\
\midrule
\textbf{avg}   & \textbf{0.30}       & \textbf{0.07}       & \textbf{5.04}       & \textbf{50.39}      & \textbf{14.07}          & \textbf{526.56}     \\
\midrule
body 0         & 0.31                & 0.10                & 5.17                & 50.39               & 14.07                   & 526.56              \\
body 1         & 0.37                & 0.06                & 4.79                & 50.39               & 14.07                   & 526.56              \\
body 2         & 0.19                & 0.06                & 5.12                & 50.39               & 14.07                   & 526.56              \\
body 3         & 0.28                & 0.10                & 5.16                & 50.39               & 14.07                   & 526.56              \\
body 4         & 0.46                & 0.05                & 5.18                & 50.39               & 14.07                   & 526.56              \\
body 5         & 0.20                & 0.07                & 5.07                & 50.39               & 14.07                   & 526.56              \\
body 6         & 0.28                & 0.09                & 4.52                & 50.39               & 14.07                   & 526.56              \\
body 7         & 0.41                & 0.06                & 5.16                & 50.39               & 14.07                   & 526.56              \\
body 8         & 0.19                & 0.07                & 5.49                & 50.39               & 14.07                   & 526.56              \\
body 9         & 0.29                & 0.09                & 5.14                & 50.39               & 14.07                   & 526.56              \\
body 10        & 0.38                & 0.07                & 5.19                & 50.39               & 14.07                   & 526.56              \\
body 11        & 0.22                & 0.06                & 5.32                & 50.39               & 14.07                   & 526.56              \\
body 12        & 0.29                & 0.11                & 4.57                & 50.39               & 14.07                   & 526.56              \\
body 13        & 0.39                & 0.06                & 4.79                & 50.39               & 14.07                   & 526.56              \\
body 14        & 0.21                & 0.05                & 5.49                & 50.39               & 14.07                   & 526.56              \\
body 15        & 0.30                & 0.10                & 4.79                & 50.39               & 14.07                   & 526.56              \\
body 16        & 0.43                & 0.06                & 4.95                & 50.39               & 14.07                   & 526.56              \\
body 17        & 0.19                & 0.05                & 4.75                & 50.39               & 14.07                   & 526.56              \\
body 18        & 0.30                & 0.11                & 5.34                & 50.39               & 14.07                   & 526.56              \\
body 19        & 0.43                & 0.06                & 4.69                & 50.39               & 14.07                   & 526.56              \\
body 20        & 0.20                & 0.06                & 4.90                & 50.39               & 14.07                   & 526.56                                 \\
\midrule
\textbf{Delta} & \textbf{0.23} & \textbf{1.45} & \textbf{0.73} & \textbf{0.78}  & \textbf{6.57}           & \textbf{5.04}   \\
\midrule
\multicolumn{6}{r}{\textbf{Actual Change:}}                                                       & {\color[HTML]{FF0000} \textbf{2.47\%}} \\
\bottomrule
\end{tabular}
\caption{Detailed comparison between the baseline and evolved agent of Fig.  \ref{fig:expBc}.}\label{table:expBc}
\end{table*}

\begin{table*}[htb]
\begin{tabular}{lllllll}
\toprule
 &	\textbf{Length} &	\textbf{Radius}	&\textbf{Density} & \textbf{Stiffness} & \textbf{Damping} & \textbf{Max Effort}\\
\midrule
\multicolumn{7}{c}{\textbf{Baseline}}                                                                                                   \\
\midrule
\textbf{avg}   & \textbf{0.66} & \textbf{0.07} & \textbf{5.00} & \textbf{50.00} & \textbf{15.00} & \textbf{20.00} \\
\midrule
body 0         & 0.30          & 0.10          & 5.00          & 50.00          & 15.00          & 20.00          \\
body 1         & 0.40          & 0.06          & 5.00          & 50.00          & 15.00          & 20.00          \\
body 2         & 0.40          & 0.06          & 5.00          & 50.00          & 15.00          & 20.00          \\
body 3         & 1.20          & 0.06          & 5.00          & 50.00          & 15.00          & 20.00          \\
body 4         & 1.20          & 0.06          & 5.00          & 50.00          & 15.00          & 20.00          \\
body 5         & 0.30          & 0.10          & 5.00          & 50.00          & 15.00          & 20.00          \\
body 6         & 0.30          & 0.10          & 5.00          & 50.00          & 15.00          & 20.00          \\
body 7         & 0.40          & 0.06          & 5.00          & 50.00          & 15.00          & 20.00          \\
body 8         & 0.40          & 0.06          & 5.00          & 50.00          & 15.00          & 20.00          \\
body 9         & 1.20          & 0.06          & 5.00          & 50.00          & 15.00          & 20.00          \\
body 10        & 1.20          & 0.06          & 5.00          & 50.00          & 15.00          & 20.00                                 \\
\midrule
\multicolumn{7}{c}{\textbf{Evolved} $\pm20\%$ (constrained torso)}                                                                                             \\
\midrule
\textbf{avg}   & \textbf{0.70} & \textbf{0.07} & \textbf{4.81} & \textbf{49.69} & \textbf{15.55} & \textbf{22.56} \\
\midrule
body 0         & 0.30          & 0.10          & 5.00          & 49.69          & 15.55          & 22.56          \\
body 1         & 0.37          & 0.05          & 5.96          & 49.69          & 15.55          & 22.56          \\
body 2         & 0.37          & 0.05          & 5.96          & 49.69          & 15.55          & 22.56          \\
body 3         & 1.44          & 0.06          & 4.77          & 49.69          & 15.55          & 22.56          \\
body 4         & 1.44          & 0.06          & 4.77          & 49.69          & 15.55          & 22.56          \\
body 5         & 0.30          & 0.10          & 5.00          & 49.69          & 15.55          & 22.56          \\
body 6         & 0.30          & 0.10          & 5.00          & 49.69          & 15.55          & 22.56          \\
body 7         & 0.46          & 0.05          & 3.46          & 49.69          & 15.55          & 22.56          \\
body 8         & 0.46          & 0.05          & 3.46          & 49.69          & 15.55          & 22.56          \\
body 9         & 1.10          & 0.05          & 4.75          & 49.69          & 15.55          & 22.56          \\
body 10        & 1.10          & 0.05          & 4.75          & 49.69          & 15.55          & 22.56                                 \\
\midrule
\textbf{Delta} & \textbf{4.55} & \textbf{7.37} & \textbf{4.01} & \textbf{0.62}  & \textbf{3.53}  & \textbf{11.33} \\
\midrule
\multicolumn{6}{r}{\textbf{Actual Change:}}                                                       & {\color[HTML]{FF0000} \textbf{5.24\%}} \\
\midrule
\multicolumn{7}{c}{\textbf{Evolved} $\pm20\%$  (constrained legs)}                                                                                                                          \\
\midrule
\textbf{avg}   & \textbf{0.67} & \textbf{0.07} & \textbf{4.98} & \textbf{50.08} & \textbf{12.44} & \textbf{17.12} \\
\midrule
ody 0         & 0.32          & 0.08          & 4.70          & 50.08          & 12.44          & 17.12          \\
body 1         & 0.40          & 0.06          & 5.00          & 50.08          & 12.44          & 17.12          \\
body 2         & 0.40          & 0.06          & 5.00          & 50.08          & 12.44          & 17.12          \\
body 3         & 1.20          & 0.06          & 5.00          & 50.08          & 12.44          & 17.12          \\
body 4         & 1.20          & 0.06          & 5.00          & 50.08          & 12.44          & 17.12          \\
body 5         & 0.25          & 0.12          & 4.42          & 50.08          & 12.44          & 17.12          \\
body 6         & 0.35          & 0.10          & 5.64          & 50.08          & 12.44          & 17.12          \\
body 7         & 0.40          & 0.06          & 5.00          & 50.08          & 12.44          & 17.12          \\
body 8         & 0.40          & 0.06          & 5.00          & 50.08          & 12.44          & 17.12          \\
body 9         & 1.20          & 0.06          & 5.00          & 50.08          & 12.44          & 17.12          \\
body 10        & 1.20          & 0.06          & 5.00          & 50.08          & 12.44          & 17.12                                 \\
\midrule
\textbf{Delta} & \textbf{0.41} & \textbf{0.24} & \textbf{0.44} & \textbf{0.17}  & \textbf{20.55} & \textbf{16.83} \\
\midrule
\multicolumn{6}{r}{\textbf{Actual Change:}}                                                       & {\color[HTML]{FF0000} \textbf{6.44\%}} \\
\bottomrule
\end{tabular}
\caption{Detailed comparison between the baseline and evolved agent of Fig.  \ref{fig:expD}.}\label{table:expD}
\end{table*}



\end{document}